\documentclass[letterpaper, 10pt, conference]{ieeeconf}  % Comment this line out if you need a4paper

\IEEEoverridecommandlockouts                              % This command is only needed if 
\usepackage{amssymb}
\usepackage{subfiles}
\usepackage{amssymb}
\usepackage{mathtools}
\usepackage{amsmath}
\let\labelindent\relax % disable labelindent IEEE conflict with enumitem
\usepackage{enumitem}
\usepackage{listings}
\usepackage{algorithm, algorithmicx}
\usepackage[noend]{algpseudocode}
\usepackage{booktabs}
\usepackage{subcaption}
\usepackage{graphicx}
\usepackage{xcolor}
\usepackage{wrapfig}
\usepackage{titlesec}
\usepackage{hyperref}
\usepackage{balance}
\usepackage{environ} % environ for toggling things on and off
\usepackage{etoolbox} % for toggling things on and off
\definecolor{algblue}{RGB}{0, 112, 192}
\usepackage{makecell}
\usepackage{xspace}
\usepackage{tabularx}
\usepackage{stfloats}  
\usepackage[backend=biber,style=ieee,natbib=true, url=false,eprint=false,doi=false]{biblatex} 
\definecolor{darkBlue}{RGB}{10,50,220}
\definecolor{customRed}{RGB}{190,110,113}
\definecolor{customGreen}{RGB}{70,170,80}

\hypersetup{
	colorlinks=true,
	linkcolor=customGreen,
	filecolor=magenta,      
	urlcolor=cyan,
	citecolor=customRed,
}
\usepackage[capitalize,noabbrev]{cleveref}

\title{\LARGE \bf\includegraphics[height=6mm]{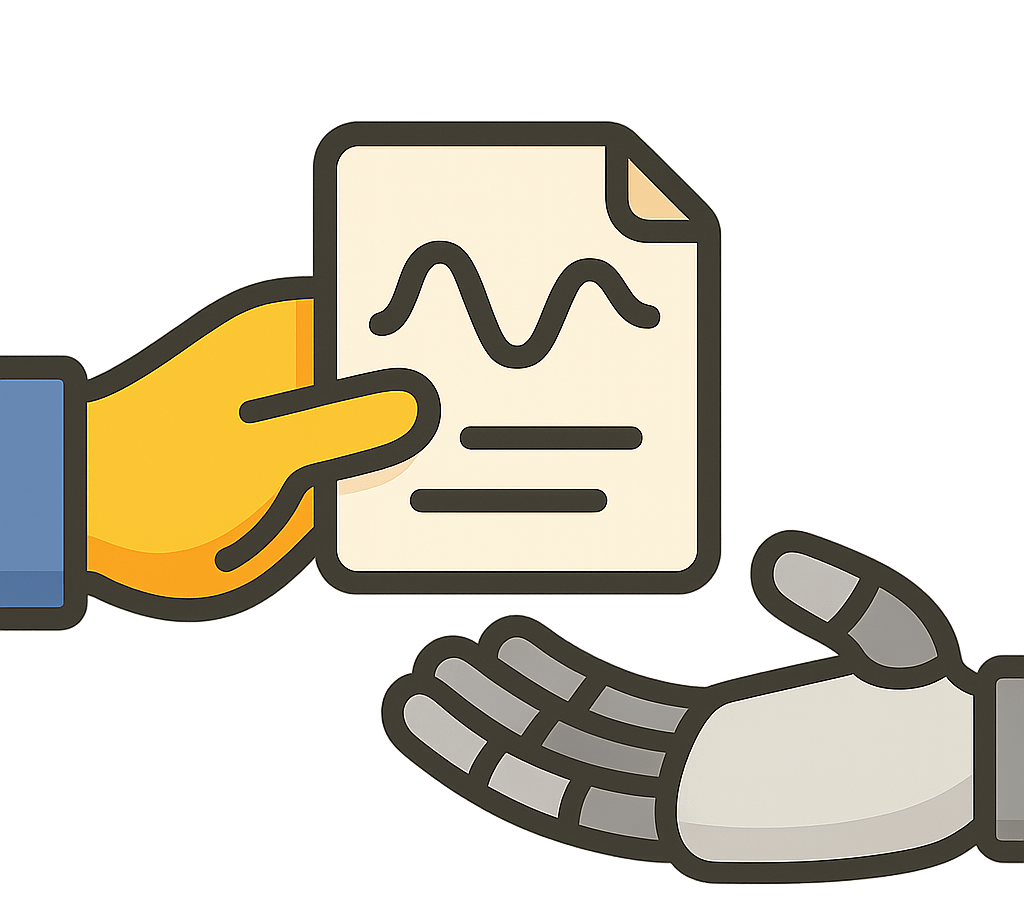}\hspace{0.2em}HAND Me the Data:\\Fast Robot Adaptation via Hand Path Retrieval
}

\newtoggle{ispreprint}
\newtoggle{isicra}
\toggletrue{isicra} % commenting this out hides icra sentences
\NewEnviron{maybePrint}[1]{%
  \iftoggle{#1}{\BODY}{}%
}

\author{
    Matthew Hong$^{\star}$, Anthony Liang$^{\star}$, Kevin Kim, Harshitha Rajaprakash, \\
    Jesse Thomason$^{\dag}$, Erdem Bıyık$^{\dag}$, Jesse Zhang$^{\dag}$ \\
    Thomas Lord Department of Computer Science,\\University of Southern California
}
\definecolor{orange}{HTML}{F28E2B}
\newcommand{\method}{HAND\xspace}

\definecolor{RoyalPurple}{HTML}{61439C}   % purple
\definecolor{ProcessBlue}{HTML}{00AFF0}   % blue
\definecolor{SeaGreen}{HTML}{3BBC9D}      % green
\newcommand{\playdata}{\mathcal{D}_\text{play}}
\newcommand{\handdata}{\mathcal{D}_\text{hand}}
\newcommand{\retrieveddata}{\mathcal{D}_\text{retrieved}}
\newcommand{\handpaths}{\{(x_t, y_t)_\text{hand}\}_{t=1}^H}
\newcommand{\playpaths}{\{(x_t, y_t)_\text{play}\}_{t=1}^T}

\newcommand{\base}{\textbf{$\pi_\text{base}$}\xspace}
\newcommand{\langc}{\texttt{Lang Cond}\xspace}
\newcommand{\clipi}{\texttt{\textcolor{SeaGreen}{CLIP (Image)}}\xspace}
\newcommand{\clipl}{\texttt{\textcolor{SeaGreen}{CLIP (Language)}}\xspace}
\newcommand{\flow}{{\texttt{\textcolor{RoyalPurple}{Flow}}}\xspace}
\newcommand{\strap}{\texttt{\textcolor{ProcessBlue}{STRAP}}\xspace}

\newcommand{\handvf}{\texttt{\textcolor{orange}{HAND(-VF)}}\xspace}
\newcommand{\handcw}{\texttt{\textcolor{orange}{HAND(-CW)}}\xspace}
\newcommand{\hand}{\texttt{\textcolor{orange}{HAND}}\xspace}
\newcommand{\task}[1]{\textsc{#1}}

\newcommand\blfootnote[1]{%
  \begingroup
  \renewcommand\thefootnote{}\footnote{#1}%
  \addtocounter{footnote}{-1}%
  \endgroup
}

\begin{document}

\makeatletter
\let\@oldmaketitle\@maketitle% Store \@maketitle
\renewcommand{\@maketitle}{\@oldmaketitle% Update \@maketitle to insert...
    \includegraphics[width=\linewidth]{sections/assets/hand_teaser_horizontal.jpg} \\[0.25em]
   \refstepcounter{figure}\footnotesize{{Fig. 1:} \method\ learns a policy from as little as 1 human hand demonstration and 4 minutes of real-world time by \emph{retrieving} from robot play data.}
  \label{fig:teaser} \medskip \vspace{-10pt}}% ... an image
\makeatother

\maketitle
\renewcommand\thefigure{\arabic{figure}}
\setcounter{figure}{1}

\thispagestyle{empty}
\pagestyle{empty}

%%%%%%%%%%%%%%%%%%%%%%%%%%%%%%%%%%%%%%%%%%%%%%%%%%%%%%%%%%%%%%%%%%%%%%%%%%%%%%%%
\begin{abstract}
We hand the community \method, a \emph{simple} and \emph{time-efficient} method for teaching robots new manipulation tasks through human hand demonstrations.
Instead of relying on task-specific robot demonstrations collected via teleoperation, \method\ uses easy-to-provide hand demonstrations to retrieve relevant behaviors from task-agnostic robot play data. 
Using a visual tracking pipeline, \method\ extracts the motion of the human hand from the hand demonstration and retrieves robot sub-trajectories in two stages: first filtering by visual similarity, then retrieving trajectories with similar behaviors to the hand.
Fine-tuning a policy on the retrieved data enables \emph{real-time learning of tasks} in under four minutes, without requiring calibrated cameras or detailed hand pose estimation.
Experiments also show that \method\ outperforms retrieval baselines by over $2 \times$ in average task success rates on real robots. 
%Videos can be found at our project website: .
% Videos can be found at our project website: \href{https://handretrieval.github.io}{https://handretrieval.github.io}. %\href{https://liralab.usc.edu/handretrieval/}{https://liralab.usc.edu/handretrieval/}.
Videos can be found at our project website: \href{https://liralab.usc.edu/handretrieval/}{https://liralab.usc.edu/handretrieval/}.
%Specifically, \method\ extracts the motion of the human hand from the provided demonstration using a visual tracking pipeline. It then retrieves sub-trajectories from the play dataset in a two-step procedure where it (1) filters by object similarity and then (2) retrieves sub-trajectories demonstrating similar motions to the human hand. By fine-tuning a policy on the retrieved robot data, \method\ enables \emph{real-time learning of tasks} without expert robot demonstrations in under \todo{XX} minutes. Unlike prior work, this approach avoids the need for calibrated cameras or detailed hand pose estimation. Experiments in simulation and on real robots additionally show that \method\ achieves \todo{XX\%} higher task success rates than retrieval baselines.
%and existing baselines by \textbf{XX\%} in task success rates with real-world learning time of less than \textbf{XX} minutes. 

% \blfootnote{$^*$Equal Contribution, $^{\dag}$ Equal Advising}
\blfootnote{$^*$Equal Contribution, $^{\dag}$ Equal Advising}
\end{abstract}

%%%%%%%%%%%%%%%%%%%%%%%%%%%%%%%%%%%%%%%%%%%%%%%%%%%%%%%%%%%%%%%%%%%%%%%%%%%%%%%%

\section{Introduction}
\label{sec:intro}
%\begin{figure}
%    \centering
%    \includegraphics[width=\linewidth]{assets/hand_teaser.png}
%    \caption{\method\ learns a policy from as little as one (1) human hand demonstration.}
%    \label{fig:teaser}
%\end{figure}
For robots to operate seamlessly in human-centric settings, they should be able to \emph{rapidly} learn new tasks with \emph{minimal human input}.
%Robots deployed in homes, warehouses, and other dynamic, human-centric settings will need to quickly learn many tasks specified by end-users. 
Achieving this goal requires robot learning algorithms that (1) scale across many tasks and (2) adapt quickly to new ones.

%To support this goal, robot learning algorithms for these settings must (scale easily across many tasks and (2) enable fast adaptation for each new task. 
Imitation learning has produced capable multi-task robot policies~\citep{team2024octo, kim2024openvla, black2024pi0, li2025hamster, geminiroboticsteam2025geminiroboticsbringingai}, but scaling is hindered by its reliance on vast amounts of expert-collected, task-specific teleoperation data~\citep{lin2025data}.
In contrast, \emph{task-agnostic play data} is far easier to collect, without requiring constant environment resets or task-specific labeling~\citep{lynch2020playdata, young2022playful, mees2022calvin}. 
The challenge is in making such unstructured data usable for teaching robots new tasks quickly. 

Therefore, we propose \method, a \emph{simple and time-efficient} approach that adapts pre-trained play policies to specific tasks using just one human hand demonstration (see \Cref{fig:teaser}). 
Unlike prior retrieval methods~\citep{nasiriany2022sailo, du2023behaviorretrievalfewshotimitation, lin2024flowretrieval, memmel2025strap, sridhar2025regent, xie2025data} that require robot demonstrations of the target task, \method\ extracts coarse, 2D relative hand motion paths from the provided human hand demonstration to guide retrieval. 
Thus, our approach enables even non-experts to teach robots without teleoperation.
%which are difficult for non-expert users to provide. 
%Instead, our key insight is to extract \emph{coarse guidance} from the hand demonstration---specifically, 2D relative hand motion paths---to retrieve diverse yet relevant behaviors from the play dataset.
%Calling back to the motivation, we aim for \method\ to be \emph{scalable} and \emph{fast}. 

\method\ enables both \emph{scalability} and \emph{speed}.
Towards \emph{scalability}, \method\ avoids the need for calibrated depth cameras~\citep{papagiannis2024rxretrievalexecutioneveryday, haldar2025point}, specialized eye-in-hand setups~\citep{kim2023giving}, or detailed hand-pose estimation~\citep{kim2023giving, lepert2025phantomtrainingrobotsrobots}. Instead, it first labels a robot play dataset with 2D gripper positions relative to the RGB camera frame by tracking the gripper using a visual point-tracking model~\citep{karaev23cotracker, karaev2025cotracker}. 
When a human hand demonstration is provided, \method\ tracks the hand trajectory with the same simple pipeline. The hand positions are then converted into 2D \emph{relative} sub-trajectories, capturing motion independent of the starting point~\citep{zhang2024extract}. 
After an initial filtering step that removes unrelated behaviors using a visual foundation model~\citep{oquab2024dinov2learningrobustvisual}, \method\ retrieves matching sub-trajectories from the play dataset based on the 2D relative hand path. 
Finally, towards \emph{speed}, a policy pre-trained on the play dataset is LoRA-fine-tuned on the retrieved sub-trajectories, encouraging the policy to specialize in the demonstrated task. Because \method\ retrieves primarily based on hand motion, it is robust to irrelevant visual features such as background clutter and lighting changes compared to purely visual retrieval methods.

Our experiments, across \textbf{10 tasks} and \textbf{550 total evaluations} in the real world on a WidowX robot demonstrate that \method\ enables quick adaptation even to long-horizon tasks, outperforming the best baseline by 3$\times$ in task completion. 
We also demonstrate that \method\ is effective with hand demonstrations collected from \emph{completely different scenes} from the robot's and across significant camera angle changes.
Finally, we perform a \emph{real-time learning} experiment, where \method\ learns a challenging long-horizon task in \textbf{under 4 minutes} of experiment time, from providing the hand demonstration to the trained policy, while being on average \textbf{5$\times$} faster to collect data for than robot teleoperation demonstrations on our WidowX arm.

\section{Related Works}
\label{sec:related}
\textbf{Robot Data Retrieval.} Prior work has demonstrated \emph{retrieval} as an effective mechanism for extracting relevant on-robot data for training robots~\citep{nasiriany2022sailo, du2023behaviorretrievalfewshotimitation, lin2024flowretrieval, memmel2025strap, sridhar2025regent, kedia2025oneshotimitationmismatchedexecution, xie2025data, sridhar2025ricladdingincontextadaptability}.
For example, SAILOR~\citep{nasiriany2022sailo} and Behavior Retrieval~\citep{du2023behaviorretrievalfewshotimitation} pre-train variational auto-encoders (VAEs) on prior robot images and actions to learn a latent embedding. This latent embedding is used to retrieve states and actions from an offline dataset similar to ones provided in expert demonstration trajectories.
However, retrieving based on learned full image encodings or even raw pixel values~\citep{sridhar2025regent} can be noisy; Flow-Retrieval~\citep{lin2024flowretrieval} instead trains a VAE to encode \emph{optical flows} indicating movement of objects and the robot arm in the scene. 
Similar to Flow-Retrieval, our method \method\ also retrieves based on robot arm movement. 
However, rather than training a dataset-specific VAE model that may not be robust to large visual differences, we retrieve from our offline robot data by primarily matching motions of a human hand demonstration using \emph{relative 2D paths} of the robot end-effector in the prior data. 
This hand path retrieval helps us robustly retrieve relevant robot arm \emph{behaviors}. %regardless of what the retrieval query videos look like.  \todo{modify claim based on final method}.

STRAP~\citep{memmel2025strap} addresses visual retrieval robustness issues of prior work by using features from DINO-v2~\citep{oquab2024dinov2learningrobustvisual}, a large pre-trained image-input foundation model for retrieval.
% , to retrieve relevant sub-trajectories. 
However, STRAP, along with all aforementioned retrieval work, assumes access to expert robot demonstrations for the target tasks.
\method, on the other hand, only requires a \emph{single}, easier-to-collect human hand demonstration that results in more \emph{time-efficient} learning of demonstrated tasks compared to methods requiring robot teleoperation data for retrieval.
Moreover, experiments demonstrate \method\ actually retrieves more task-relevant trajectories and therefore attains higher success rates compared to these methods. 

\textbf{Learning From Human Hands.}
Similar to \method, a separate line of work has proposed methods to use human hands to learn robot policies.
One approach is to train models on human video datasets to predict future object flows~\citep{xu2024flow,yuan2024general} or human affordances~\citep{bahl2023affordances, kuang2024ram}.
These intermediate affordance and flow representations are then used to either train a policy conditioned on this representation~\citep{xu2024flow} on robot data or control a heuristic policy~\citep{yuan2024general, bahl2023affordances, kuang2024ram}.
%We are instead focused on retrieving from \emph{robot play data}, containing ground truth robot action data, allowing us to train arbitrarily expressive imitation learning policies on the data.
Other works focus on learning directly from human hands~\citep{papagiannis2024rxretrievalexecutioneveryday, haldar2025point, kim2023giving, kareer2024egomimicscalingimitationlearning, lepert2025phantomtrainingrobotsrobots}. 
These works generally use hand-pose detection models aided by multiple cameras or calibrated depth cameras to convert hand poses directly
% , obtained either from provided demonstrations or retrieved human data, 
to robot gripper keypoints~\citep{papagiannis2024rxretrievalexecutioneveryday, haldar2025point, lepert2025phantomtrainingrobotsrobots}.
However, works that exclusively retrieve human data are restricted to constrained policy representations as they must match human hand poses to robot gripper poses.
\citet{kim2023giving} instead use an eye-in-hand camera mounted on a human demonstrator's forearm to train an imitation learning policy conditioned on robot eye-in-hand camera observations.
Unlike these prior works, \method\ only requires a single RGB camera from which the robot gripper can be seen.
% instead of calibrated RGBD cameras and hand-pose detection models~\citep{papagiannis2024rxretrievalexecutioneveryday, haldar2025point, lepert2025phantomtrainingrobotsrobots} or specific eye-in-hand observations~\citep{kim2023giving}. 
Also, we focus on retrieving robot play data, allowing us to train arbitrarily expressive policies without constrained policy representations~\citep{papagiannis2024rxretrievalexecutioneveryday, haldar2025point, lepert2025phantomtrainingrobotsrobots} or intermediate representations~\citep{xu2024flow, yuan2024general, bahl2023affordances, kuang2024ram}.

\section{\method: Fast Robot Adaptation\\via Hand Path Retrieval}
\label{sec:method}
\begin{figure*}[t]
    \centering
    \includegraphics[width=\linewidth]{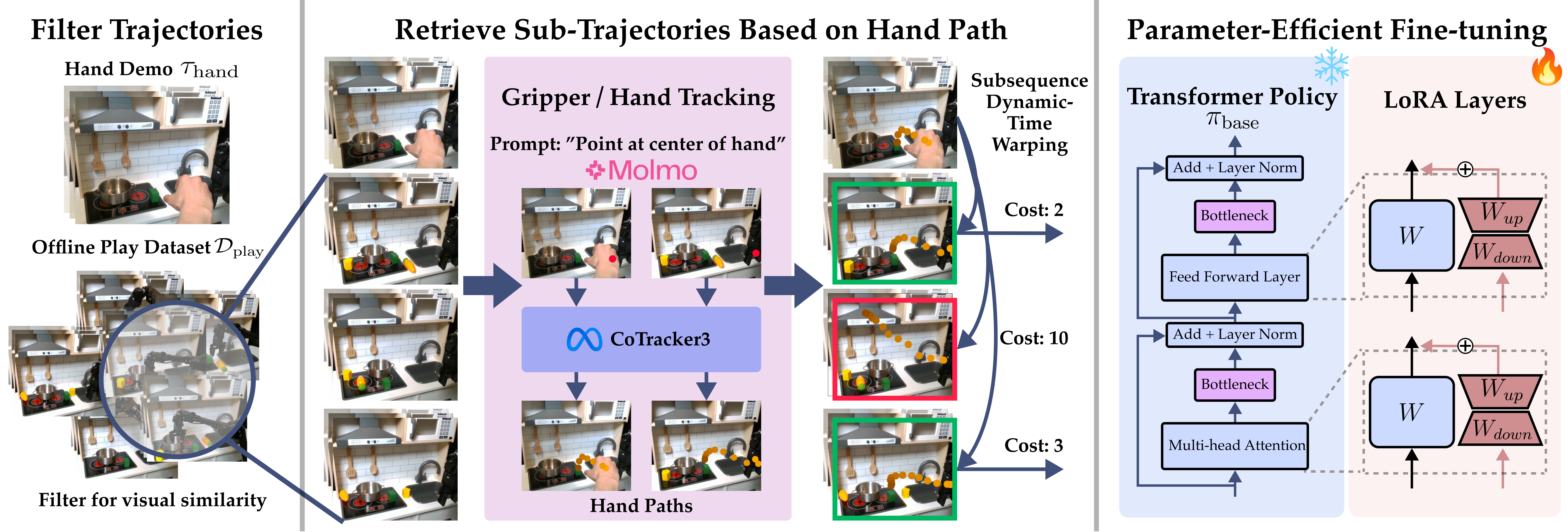}
    \caption{\textbf{HAND} enables fast-adaptation to a new target task by using an easy-to-provide hand demonstration of the target task (Left). We propose a two-step retrieval procedure where we first filter the trajectories in the offline play dataset, $\playdata$, for visually similar trajectories based on features from a pretrained vision model. We use off-the-shelf, pretrained hand detection and point tracking to construct 2D paths of the motion for both the human hand and robot end-effector. We use these paths as a distance metric to retrieve relevant trajectories from the play dataset (Middle) for quickly fine-tuning a pretrained transformer policy on the target task (Right).}
    \label{fig:method}
\end{figure*}

%\subsection{Problem Formulation: Retrieving Relevant Sub-Trajectories with Hand Demos}
%\paragraph{Problem Formulation.} 
\subsection{Preliminaries and Formulation.}
\label{sec:method:preliminaries}
We assume access to a dataset of task-agnostic robot play data, $\playdata$, consisting of trajectories ${\tau_i = \{(o_t, a_t)\}_{i=1}^T}$, where each $o_t$ is per-timestep observation that includes RGB images of the robot gripper and robot proprioceptive information, and $a_t$ is the robot action. 
These trajectories may span many scenes, tasks, and time horizons.
We do not assume task labels (e.g., language labels), as data collection is easier to scale without labeling each sub-trajectory in a long-horizon play trajectory.\footnote{\Cref{sec:experiments} demonstrates that \method can also incorporate language labels as extra policy conditioning.}
%We assume the RGB camera's angle relative to the robot base is similar across trajectories,\footnote{Our experiments demonstrate \method works even under significant camera angle shift} such as with a mobile manipulator with a fixed head camera.
% or a mobile manipulation robot with a fixed head camera relative to its arm. 

In contrast to retrieval methods that rely on robot demonstrations for each target task~\citep{nasiriany2022sailo, du2023behaviorretrievalfewshotimitation, lin2024flowretrieval, memmel2025strap}, we assume access to easy-to-provide human hand demonstrations. %\footnote{\method also outperforms baselines even when they have access to robot demonstrations.} 
For each task, a human records their hand movement without teleoperating the robot. On our real-world setup, these hand demonstrations, $\handdata$, are on average $5\times$ faster to collect than robot teleoperation data. 
Moreover, producing high-quality hand demonstrations typically requires far less effort than robot teleoperation~\citep{hand_vs_robto_mdpi_2025, li2025trainrobotsimpactdemonstration}. Each video in $\handdata$ consists of a sequence of RGB images $o_1, \ldots, o_H$, captured from a similar viewpoint relative to the human hand as trajectories in the robot play data relative to the robot gripper.\footnote{\Cref{sec:experiments} demonstrates \method works under large camera angle shifts.}

%In contrast to prior methods for retrieving robot data that assume robot demonstrations for the target task~\citep{nasiriany2022sailo, du2023behaviorretrievalfewshotimitation, lin2024flowretrieval, memmel2025strap}, we assume access to easy-to-provide human hand demonstrations.\footnote{Our experiments show that \method also outperforms baselines when they have access to robot demos.}
%Specifically, for each desired target task, we assume a human demonstrates it by video recording their hand movement \emph{without} teleoperating the robot. 
%Our experiments show that these human hand demonstrations, $\handdata$, can be collected on average 5$\times$ faster than robot teloperation demonstrations on our real-world manipulation setup. 
%Additionally, high-quality human hand demonstrations generally require significantly less effort than high-quality robot teleoperation demonstrations~\citep{hand_vs_robto_mdpi_2025, li2025trainrobotsimpactdemonstration}.
%Each demonstration video in $\handdata$ consists of a sequence of RGB image observations $o_1,\ldots, o_H$.
%We assume that the hand videos are captured from approximately the same position relative to the human hand as the robot play data's image observations to the robot gripper.

Given $\playdata$ and $\handdata$, we aim to train a policy $\pi_{\theta}(a \mid o)$ to perform the target task demonstrated by the human in $\handdata$.
Since we do not assume task labels in $\playdata$ and we are provided no expert robot teleoperation demonstrations, we must \emph{retrieve} sub-trajectories indicating how to perform the behavior demonstrated in $\handdata$ from $\playdata$ for training $\pi$.
We denote this retrieved dataset, later used for imitation learning, as $\retrieveddata$. 
Moreover, following our motivation in \Cref{sec:intro}, we aim for our method to be \emph{fast}, so that non-expert end-users can easily train the robot for many downstream tasks.

Thus, the key challenges we resolve in our method \method are: (1) designing a representation that can unify the behaviors in robot sub-trajectories and human hand demonstrations (\Cref{sec:method:metric}), (2) retrieving relevant sub-trajectories based on a suitable distance metric between these representations (\Cref{sec:method:retrieval}), and (3) quickly training a policy that can perform various unseen target tasks with a high success rate without expert demonstrations (\Cref{sec:method:training}). 
\begin{maybePrint}{isicra}
See \Cref{fig:method} for an overview.
\end{maybePrint}
\begin{maybePrint}{ispreprint}
See \Cref{fig:method} for an overview and \Cref{alg:hands} for full algorithm pseudocode.
\end{maybePrint}

\subsection{Path Distance as a Unifying Representation for Retrieval}
\label{sec:method:metric}
Prior robot retrieval methods assume access to expert demonstrations from which they extract proprioceptive information (e.g., joint angles and actions) alongside visual features for retrieval~\citep{nasiriany2022sailo, du2023behaviorretrievalfewshotimitation, lin2024flowretrieval, memmel2025strap, sridhar2025regent}. However, since $\handdata$ contains only visual data and no robot actions, retrieval based purely on appearance can be noisy---especially due to the visual domain gap between hand demonstrations in $\handdata$ and robot demonstrations in $\playdata$ (c.f., \Cref{fig:method}, left). To address these issues, we propose an embodiment-agnostic, behavior-centric retrieval metric that enables matching between $\handdata$ and $\playdata$ based on demonstrated behaviors rather than appearance.

\textbf{Using 2D Paths for Retrieval.}  
The movement of the robot end-effector over time provides rich information about its behavior~\citep{li2025hamster}. 
We represent behaviors in both datasets using the paths traced by the human hand or the gripper. 
Because we assume access only to an RGB camera from which the hand or the gripper is visible (i.e., no depth), we construct these paths in 2D relative to the camera viewpoint for both $\playdata$ and $\handdata$.\footnote{If both datasets have additional calibrated depth information, \method can also operate on 3D paths.}

\textbf{Obtaining Paths from Data.}  
To extract paths, we use CoTracker3~\citep{karaev2025cotracker}, an off-the-shelf point tracker capable of tracking 2D points across video sequences, even under occlusion. 
CoTracker3 only requires a single point on the gripper or hand to generate a complete trajectory. 
We use Molmo-7B~\citep{molmo2024}, an open-source 7B image-to-point foundation model, to automatically select this point by prompting it at the \emph{midpoint} of each trajectory with either ``Point at the center of the hand'' or ``Point to the robot gripper.'' 
Using the middle frame ensures a higher chance of visibility in case the gripper or hand is not yet in frame at the beginning or occluded at the end.\footnote{Points can also be obtained heuristically, e.g., if the robot starts from the same position in each $\playdata$ traj.}

Given the 2D point $(x, y)_\text{hand}$ or $(x, y)_\text{play}$ from the middle frame, we use CoTracker3 to perform bidirectional point tracking, resulting in a 2D path $p_{\text{hand}} = \handpaths$ or $p_{\text{play}}=\playpaths$ for each trajectory. See the \textcolor[HTML]{EEB5EC}{Gripper/Hand Tracking} block of \Cref{fig:method} for a visualization of this pipeline. 
Next, we describe how we use 2D paths to retrieve sub-trajectories from $\playdata$.

\subsection{Retrieving Relevant Sub-Trajectories using Path Distance}
\label{sec:method:retrieval}

\textbf{Background.} 
For identifying relevant sub-trajectories in $\playdata$, we use Subsequence Dynamic Time Warping (S-DTW)~\citep{muller2021fundamentals}, an algorithm for aligning a shorter sequence to a portion of a longer reference sequence prior work has demonstrated effective for sub-trajectory retrieval~\citep{memmel2025strap}. 
Given a query sequence $Q=\{q_1, q_2, \dots, q_H\}$ and a longer reference sequence $R=\{r_1, r_2, \dots, r_T\}$, where $T > H$, the goal of S-DTW is to find a contiguous subsequence of $R$ that minimizes the total cumulative distance between elements of both sequences.
In \method, the query sequences are the 2D hand demonstration paths $\handpaths$ and the reference sequences are the 2D paths generated from long-horizon robot play data $\playpaths$.
% detailed in \Cref{sec:method:metric}.

\textbf{Sub-Trajectory Preprocessing.} 
To preprocess the datasets for S-DTW, we first segment %the hand demonstrations, $\handdata$, and 
the offline play dataset, $\playdata$, into variable-length sub-trajectories using a simple heuristic based on proprioception proposed in several prior works~\citep{shridhar2023perceiver, memmel2025strap}.
In particular, we split the trajectories whenever the acceleration or velocity magnitude (depending on what proprioception data is available) drops below a predefined $\epsilon$ value, corresponding to when the teleoperator switches between tasks.
We find that this simple heuristic can reasonably segment trajectories into atomic components resembling lower-level primitives.
We also split the hand demonstrations evenly into smaller sub-trajectories based on how many subtasks the human operator determined they have completed.
After sub-trajectory splitting, we have two sub-trajectory datasets, ${\mathcal{T}_{\text{hand}} = \{t_{1:a}^{i}, t_{a:b}^{i}, \dots, t_{H_i - |p_{\text{hand}}^i|:H_i}^i \forall \, \tau^{i}_\text{hand} \in \handdata \}}$ and ${\mathcal{T}_{\text{play}} = \{t_{1:a}^{j}, t_{a:b}^{j}, \dots,  t_{T_j-|p_{\text{play}}^j|:T}^j \forall \, \tau^{j}_\text{play} \in \playdata \}}$ where $|p_\text{hand}^i|$ and $|p_\text{play}^j|$ are the lengths of the last sub-trajectory paths of trajectories $i, j$ from $\handdata$ and $\playdata$, respectively.
Finally, each sub-trajectory is represented in \emph{relative 2D coordinates}, i.e., ${p_t = [x_{t+1} - x_t, y_{t+1} - y_t]}$.
%prior work that proposes to cluster trajectories based on relative embedding differences~\citep{zhang2024extract}, each sub-trajectory is represented in \emph{relative 2D coordinates}, i.e., ${p_t = [x_{t+1} - x_t, y_{t+1} - y_t]}$.
%where each 2D point is subtracted by the previous timestep point, effectively reducing the length of the segmented sub-trajectories by 1.
Relative coordinates ensure retrieval invariance to the initial positions of the hand or gripper~\citep{zhang2024extract}.
%encouraging robot motions, e.g., picking or sliding motions, to be retrieved regardless of the starting positions.

\textbf{Visual Filtering.} 
One issue with retrieving sub-trajectories based only on path distance is that different tasks can have similar movement patterns. For example, tasks like ``pick up the mug'' and ``pick up the cube'' can appear nearly identical in 2D path space~\citep{li2025hamster}. 
But, the retrieved trajectories for one task may not benefit learning of the other;
%This similarity can confuse the robot if it tries to perform ``pick up the mug'' but retrieves a trajectory intended for ``pick up the cube.''
%One problem with directly retrieving sub-trajectories using only path distance is that different tasks may share similar paths. 
%For example, ``pick up the mug'' and ``pick up the cube'' can look similar in 2D path space.
%Yet, if we are trying to have the robot ``pick up the mug,'' then it may be confused by a retrieved ``pick up the cube'' trajectory.
since we do not assume task labels in $\playdata$, a policy directly trained on ``pick up the cube'' retrieved sub-trajectories may still fail to pick up a mug.
Therefore, before retrieving sub-trajectories with paths, we first run a visual filtering step to ensure that the sub-trajectories we retrieve will be task-relevant. 
We use an object-centric visual foundation model, namely DINOv2~\citep{oquab2024dinov2learningrobustvisual}, to first filter out sub-trajectories performing unrelated tasks with different objects.
Specifically, we use the DINOv2 first and final frame embedding differences, representing visual object movement from the first to last frame, between human hand demonstrations and robot play data to filter $\mathcal{T}_\text{play}$.
We find that using this simple method is sufficient to filter out most irrelevant sub-trajectories.
%and argue that DTW should be used for aligning motions rather than visual embeddings
For a given image sequence $o_{1:H}^\text{hand}$ from a hand sub-trajectory and image sequence $o_{1:T}^\text{play}$ from a robot play sub-trajectory, we define the cost as:
\begin{align}
    \text{C}_\text{visual}(o_{1:H}^{\text{hand}}, o_{1:T}^{\text{play}}) &= \underbrace{||\text{DINO}(o_1^\text{hand}) - \text{DINO}( o_1^\text{play})||_2^2}_{\text{first frame DINO embedding difference}} \nonumber \\
    &+ \underbrace{||\text{DINO}(o_H^\text{hand}) - \text{DINO}(o_T^\text{play})||_2^2}_{\text{last frame DINO embedding difference}}.
\label{eq:visual_filtering}
\end{align}
We take the $M$ trajectories with lowest cost as possible retrieval trajectories from $\playdata$ for each human demonstration sub-trajectory in $\mathcal{T}_\text{hand}$.
The rest are ignored for those hand demonstrations.

\textbf{Retrieving Sub-Trajectories.}
Finally, we then employ S-DTW to match the target sub-trajectories, $\mathcal{T}_{\text{hand}}$, to the set of visually filtered segments $ \in \mathcal{T}_{\text{play}}$. 
Given two sub-trajectories, $t_i \in \mathcal{T}_{\text{play}}$ and $t_j \in \mathcal{T}_{\text{hand}}$, S-DTW returns the cost along with the start and end indices of the subsequence in $t_j$ that minimizes the path cost (see \Cref{fig:method}).
We select the $K$ matches from $\playdata$ with the lowest cost to construct our retrieval dataset, $\retrieveddata$.

\subsection{Putting it All Together: Fast-Adaptation with Parameter-Efficient Policy Fine-tuning}
\label{sec:method:training}
We aim to enable fast, data-efficient learning of the task demonstrated in $\handdata$. 
To this end, we first pretrain a task-agnostic base policy $\pi_{\text{base}}$ on $\playdata$ with standard behavior cloning (BC) loss.
While our approach is compatible with any policy architecture, we use action-chunked transformer policies~\citep{zhao2023learning} due to their suitability for low-parameter fine-tuning and strong performance in long-horizon imitation learning~\citep{zhao23aloha, zhao2024alohaunleashedsimplerecipe, haldar2024baku, black2024pi0}. 

% \begin{wrapfigure}{r}{0.5\textwidth}
% \vspace{-1em}
% \begin{minipage}{0.5\textwidth}
% \begin{algorithm}[H]
% \caption{\textsc{HAND Pseudocode}}
% \begin{algorithmic}[1]
% \Require $\mathcal{D}_\text{hand}$, $\mathcal{D}_\text{play}$, \# visual-filtered $M$, \# retrieved $K$
% \State Train $\pi_\text{base}$ on $\mathcal{D}_\text{play}$ via BC
% \State Segment $\mathcal{D}_\text{play}$, $\mathcal{D}_\text{hand}$ into $\mathcal{T}_\text{play}$, $\mathcal{T}_\text{hand}$ 
% \For{$\tau^\text{hand} \in \mathcal{T}_\text{hand}$}
%     \State Filter top-$M$ from $\tau^\text{play}$ w/ DINO-v2
%     \For{each filtered $\tau^\text{play}$}
%         \State Track 2D hand paths (CoTracker3)
%         \State Retrieve $K$ best via S-DTW on paths
%     \EndFor
% \EndFor
% \State Fine-tune $\pi_\text{base}$ w/ adapters on $\retrieveddata$ $\to$ $\pi_\theta$
% \State \Return $\pi_\theta$
% \end{algorithmic}
% \label{alg:hand_summary}
% \end{algorithm}
% \end{minipage}
% \vspace{-1em}
% \end{wrapfigure}
\textbf{Adapting to $\retrieveddata$.} To rapidly adapt to a new task with minimal data, we leverage parameter-efficient fine-tuning using \emph{task-specific adapters}—small trainable modules that modulate the behavior of the frozen base policy. 
Adapter-based methods have shown promise in few-shot imitation learning~\citep{liang2022transformer, liu2024tail}, making them ideal for our limited retrieved dataset $\retrieveddata$.
%Following the findings of \citet{liu2024tail}, 
Specifically, we insert LoRA layers~\citep{hu2022lora} into the transformer blocks of $\pi_{\text{base}}$. 
These are low-rank trainable matrices (about $0.1\%$–$2\%$ of $\pi_{\text{base}}$'s parameters) inserted between the attention and feedforward layers (see \Cref{fig:method}, \textcolor[HTML]{E99999}{LoRA Layers}). 
During fine-tuning, we update only the parameters of these LoRA layers, $\theta$, using $\retrieveddata$.
%At deployment time, this setup enables a plug-and-play interface: task-specific adapters can be dynamically swapped in to perform different tasks without retraining the base policy. We fine-tune using a simple behavior cloning (BC) objective.

\textbf{Loss Re-Weighting.} While our retrieval mechanism identifies sub-trajectories relevant to the target task, not all will be equally useful. 
Following prior work~\citep{sridhar2025regent, xie2025data, sridhar2025ricladdingincontextadaptability}, 
we reweight the BC loss with an exponential term $\in (0, \infty)$ (similar to AWR~\citep{peng2019advantage}), where each sub-trajectory is weighted based on its S-DTW similarity to the hand demonstration.
Intuitively, this upweights the loss of the most relevant examples in $\retrieveddata$ and downweights those that are less relevant. 
%We exponentiate and clip the similarity-derived costs to ensure stability and control the range of influence. 
Finally, because trajectory cost scales vary depending on the task being retrieved and the features being used for S-DTW, we rescale the S-DTW costs $C_{i, \text{path}}$ to a fixed range. For each $\tau_i \in \retrieveddata$, its weight $e^{-C_{i, \text{path}}}$ is scaled to between $[0.01, 100]$, where the normalization term comes from the sum of costs of all trajectories in $\retrieveddata$.
Let the normalized weight for a trajectory be $w_i = \exp(-C_{i, \text{path}})$ and the behavioral cloning loss be $L_i(a,o) = -\log \pi_{\theta}(a \mid o)$. The total loss is then the weighted average over the dataset $\mathcal{D}$:
\begin{equation}
    \mathcal{L}_{\text{BC};\theta} = \frac{1}{|\retrieveddata|}\sum_{\tau_i \in \retrieveddata} w_i \times L_i(a,o).
\label{eq:bc_loss}
\end{equation}
We summarize \method in the pseudocode in \Cref{alg:hand_summary}. 
\begin{maybePrint}{ispreprint}
For the full algorithm, see \Cref{alg:hands}; for full implementation details and hyperparameters, see \Cref{sec:appendix:impl_details}.
\end{maybePrint}

\begin{algorithm}[t]
\caption{\textsc{HAND Pseudocode}}
\begin{algorithmic}[1]
\Require $\mathcal{D}_\text{hand}$, $\mathcal{D}_\text{play}$, threshold $\epsilon$, \# visual-filtered trajectories $M$, \# retrieved sub-trajectories $K$
\State Train base policy $\pi_\text{base}$ on $\mathcal{D}_\text{play}$ via behavior cloning
\State Segment both $\playdata$ and $\handdata$ into sub-trajectory datasets w/ threshold $\epsilon$: $\mathcal{T}_\text{play}$, $\mathcal{T}_\text{hand}$
%\State $\mathcal{T}_\text{hand}, \mathcal{T}_\text{play} \gets \texttt{SubTrajSegmentation}(\handdata, \epsilon), \texttt{SubTrajSegmentation}(\playdata, \epsilon)$
\For{$\tau^\text{hand} \in \mathcal{T}_\text{hand}$}
    \State Filter top-$M$ visually similar $\tau^\text{play} \in \mathcal{T}_\text{play}$ via DINO-based $\text{C}_\text{visual}$
    \For{each filtered $\tau^\text{play}$}
        \State Track 2D hand paths with Molmo + CoTracker3
        \State Retrieve $K$ best-matching segments via S-DTW on relative path similarity
    \EndFor
\EndFor
\State Fine-tune $\pi_\text{base}$ on retrieved data with adapter layers $\theta$ to obtain $\pi_\theta$ with $\mathcal{L}_{\text{BC};\theta}$ \cref{eq:bc_loss}
\State \Return $\pi_\theta$
\end{algorithmic}
\label{alg:hand_summary}
\end{algorithm}

\section{Experiments}
\label{sec:experiments}
Our experiments demonstrate the efficacy of \method as a robot data retrieval pipeline and evaluate its ability to quickly learn to solve downstream tasks.
To this end, we organize our experiments to answer the following questions, in order:

\begin{enumerate}[label=\textbf{(Q\arabic*)}  ]
    \item \label{q1} How well can \method retrieve \emph{task-relevant} behaviors?
    \item \label{q2} Does \method support hand demonstrations from \emph{unseen scenes} and is it \textit{robust} to visual shifts?
    \item \label{q3} How does \method perform in policy learning?
    \item \label{q4} Can \method enable \emph{real-time} adaptation?
    %\item \label{q5} How can \method quickly learn long-horizon tasks with humans-in-the-loop?
\end{enumerate}

\subsection{Experimental Setup} 
\begin{figure}[t]
    \centering
    \includegraphics[width=0.8\linewidth]{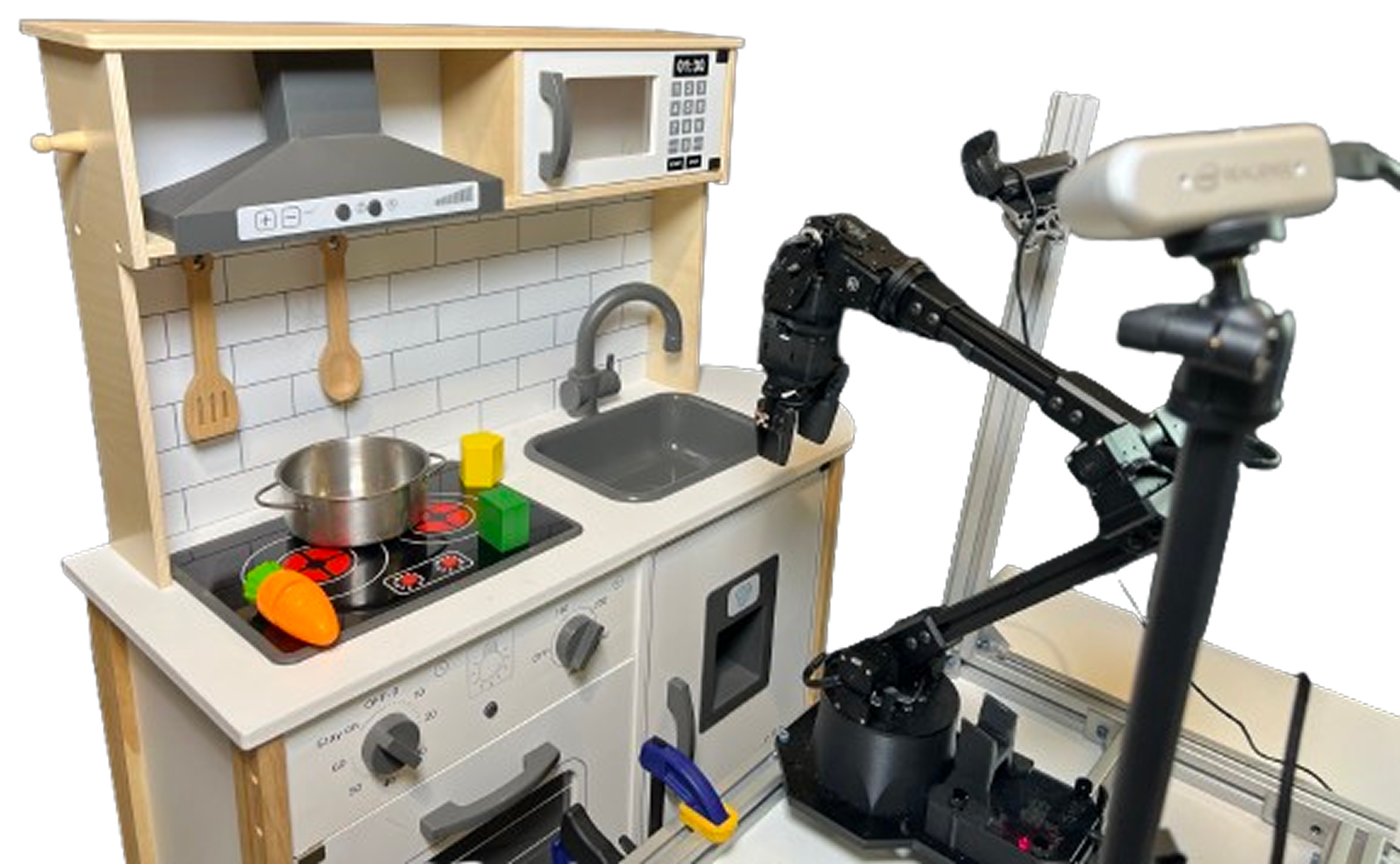}
    \caption{\textbf{WidowX Robot Arm Setup.} We evaluate the scalability of HAND on 10 manipulation tasks on a WidowX robot arm in a kitchen setup~\citep{walke2023bridgedata}.}
    \label{fig:widowx}
\end{figure}

We evaluate HAND on a real-world multi-task kitchen environment using the WidowX robot arm.
Our robot environment setup is shown in \Cref{fig:widowx}. 
We use an Intel Realsense D435 camera
as an external camera and a Logitech C920 as an over-the-shoulder camera.

\textbf{Evaluation Tasks:} We first evaluate on 10 total tasks. 
We first evaluate on three standard tasks: \task{Reach Green Block}, \task{Press Button}, and \task{Close Microwave}.
Then, we introduce three challenging long-horizon tasks: \task{Put K-Cup in Coffee Machine}, \task{Blend Carrot}, and \task{Cook Carrot}, which demand high precision and span more than 150 timesteps at a 5 Hz control frequency. In particular, \task{Cook Carrot} is composed of four shorter tasks, \task{Slide Pot} $\rightarrow$ \task{Put Object in Pot} $\rightarrow$ \task{Put Lid on Pot} $\rightarrow$ \task{Turn Stove Knob}, including non-prehensile tasks (e.g., slide pot) and taking $\sim300$ steps to complete even for expert teleoperators.
For our long-horizon tasks, we provide one hand demonstration to perform retrieval for each subtask. 
These tasks highlight the ability of \method to acquire and execute complex behaviors in real time. 
Partial success is provided for tasks composed of multiple subtasks.

\textbf{Play Dataset Collection:} % We demonstrate that \method can also scale to real-world scenarios by evaluating on several manipulation tasks in a kitchen setup shown in \Cref{fig:widowx}.
% We collect a task-agnostic play dataset of about $50$k transitions. 
We collect a task-agnostic play dataset containing a total of 50k transitions, each trajectory having an average of 230 timesteps and covering multiple tasks, collected at 5hz.
The full dataset required roughly four hours to collect. 
We place distractor objects not used in target tasks in the environment that teleoperators interact with during play data collection to ensure the play data does not mirror evaluated tasks.
The dataset is split into two, with about 1 hour corresponding to the scene for \task{Cook Carrot}.
To evaluate language-conditioned methods, we manually annotate the \task{Cook Carrot} scene of the dataset with language, which takes an additional 87 minutes.
% Human teleoperators were instructed to freely interact with the available objects in the scene without being bound to specific task goals. 
During data collection and evaluation, movable task objects are randomized in a 5’’ x 7’’ region within the workspace.

% \begin{figure}[t]
%     \centering
%     \includegraphics[width=\linewidth]{assets/calvin_results.png}
%     \caption{\textbf{CALVIN Results.} Task success rate of \hand and baseline methods on the CALVIN \texttt{ABC-D} task across three random seeds. Ablations of \hand are denoted by hatches. \hand and ablations outperform the next best baseline \flow on task success rate across all tasks.}
%     \label{fig:calvin results} 
%     % \textcolor{red}{TODO: fix colors to correspond with method colors}
% \end{figure}

% \textbf{CALVIN} contains unstructured, teleoperated play data in four tabletop manipulation environments \{\texttt{A},\texttt{B},\texttt{C},\texttt{D}\}, that share the same set of objects, but have different visual textures and static object locations (e.g., slider, button, switch), shown in {\Cref{fig:environment} (Left)}. 
% Because it is infeasible to provide explicit human hand demonstrations in CALVIN, we instead perform end-effector point-tracking on expert task demonstrations to mimic the effect of hand-based tracking. %experiments using a few expert task demonstrations following \citet{memmel2025strap}.
% CALVIN offers four unique environments, 
% We uniformly sample $N=6$ task-specific expert trajectories from environment \texttt{D} as $\handdata$, and utilize about $17$k trajectories from environments \{\texttt{A},\texttt{B},\texttt{C}\} as $\playdata$.
% We evaluate our fine-tuned policy in environment \texttt{D} across 3 tasks.

% \textbf{LIBERO:}

\textbf{Baselines:} We compare \hand to the following baselines: % and ablations. 
% We consider the following baseline methods:
\begin{itemize}
\item \base: the base policy pre-trained on play data;
%\item \ft\ fine-tunes the base policy on target task expert robot demonstrations if available;
\item \langc: \base with language-conditioning;
\item \clipl: retrieves based on cosine similarity between language embedding of target task (rather than hand demo) and language embedding of the play data;
\item \clipi: retrieves based on cosine similarity between language embedding of target task and image embedding of the play data;
\item \flow~\citep{lin2024flowretrieval}: trains a VAE on pre-computed optical flows for $\playdata$ from GMFlow~\citep{xu2022gmflow} and retrieves individual states-action pairs based on latent motion similarity; and
\item \strap~\citep{memmel2025strap}: also uses S-DTW for sub-trajectory retrieval but uses S-DTW distance based solely on Euclidean distance between pre-trained DINO-v2 image embeddings.
%\item \handee\ uses ground truth 3D pose information (not assumed by \method generally) to perform retrieval;
%\item \handcw\ uses uniform trajectory cost weights for $\mathcal{L}_\text{BC}$ in place of $\exp(-C_{i, \text{path}})$ (\cref{eq:bc_loss});
%\item \handvf does not apply DINO-based visual filtering for retrieval;
\end{itemize}
\strap and \flow assume access to expert \emph{robot} demonstrations for both retrieval and fine-tuning. 
In our setting, we do not assume such demonstrations, therefore, unless otherwise noted, we adopt them without expert fine-tuning.
While \strap and \flow originally propose training policies from scratch, we instead apply LoRA fine-tuning—as with \hand—which we found to yield better performance for these baselines.

\textbf{Policy Architecture:} To ensure fair comparison, all methods use a three-layer action-chunking transformer (similar to ACT~\citep{zhao2023learning}) decoder policy where applicable.
The input to the transformer policy is a sequence of image tokens corresponding to the external and over-shoulder camera views. 
Conditioned on the current image observation, the model predicts an action chunk corresponding to a second of execution.
\begin{maybePrint}{ispreprint}
We refer the reader to \Cref{sec:appendix:impl_details} for implementation details and \Cref{sec:appendix:calvin_results} for extensive ablation results.
\end{maybePrint}

\begin{table}[t]
\centering
\resizebox{\linewidth}{!}{%
\begin{tabular}{lccc}
\toprule
 & Reach Green Block & Push Button & Close Microwave \\
\midrule
\flow & 7/25 & 0/25 & 0/25 \\
\strap & 5/25 & 0/25 & 2/25\\
\handvf & 9/25 & 13/25 & 9/25 \\
\hand & \textbf{15/25} & \textbf{18/25} & \textbf{11/25} \\
\bottomrule
\end{tabular}
}
\caption[\hand retrieves more sub-trajectories performing the task.]{\textbf{Number of retrieved sub-trajectories performing demonstrated task.} \hand retrieves more task-performing sub-trajectories than \flow and \strap.}
\label{tab:retrieved_subtraj_comparison}
\end{table}

\subsection{Experimental Evaluation}

% \begin{wraptable}{R}{0.5\textwidth}
% \centering
% \vspace{-1em}
% \begin{tabular}{l|@{\hspace{8pt}}c@{\hspace{8pt}}@{\hspace{8pt}}c@{\hspace{8pt}}c}
% \toprule
%  & Block & Button & Microwave \\
% \midrule
% STRAP & 12/25 & 23/25 & 18/25\\
% \method & \textbf{20/25} & \textbf{25/25} & \textbf{24/25} \\
% \bottomrule
% \end{tabular}
% \caption{\textbf{Comparison of retrieved trajectories.} \method retrieves more task relevant trajectories out of the top K=25 matches compared to STRAP.}
% \label{tab:retrieved_subtraj_comparison}
% \end{wraptable}
%\footnotetext[label=fn:flow]{For \flow, we examine the top retrieved states.}
%\footnotetext[1]{For \flow, we examine the top retrieved states.}

\textbf{\ref{q1}: \method retrieves more task-relevant data.} 
We analyze the quality of retrieved sub-trajectories between \flow, \strap, and \hand. \strap and \hand both use S-DTW-based trajectory retrieval, but \strap relies purely on visual DINO-v2 embeddings for retrieval.
We provide a single hand demonstration of three real robot tasks and retrieve the top $K=25$ matches from $\playdata$.
Compared to \strap and \flow, we observe in \Cref{tab:retrieved_subtraj_comparison} that \hand retrieves more trajectories in which the robot performs the demonstrated task.
\strap relies exclusively on visual similarity, while \flow relies exclusively on motion similarity. Both methods struggle when there is a significant visual or motion gap between the target demonstrations (e.g., human hand videos) and the play dataset.
In particular, for the Push Button task, \strap is unable to retrieve any relevant trajectories in its top matches.

We also observe that \textbf{visual filtering is necessary} to retrieve trajectories where the target object is interacted with, as demonstrated by \handvf, an ablation of \hand without visual filtering (\Cref{sec:method:retrieval}), having 30\% worse retrieval performance than \hand in \Cref{tab:retrieved_subtraj_comparison}.
%Furthermore, \method's visual filtering step allows it to retrieve $+30\%$ more relevant trajectories across all tasks.
% We provide qualitative comparisons of retrieved trajectories by in \Cref{sec:appendix:qualitative_retrieval_analysis}.
\begin{figure}[t]
    \centering
    \includegraphics[width=\linewidth]{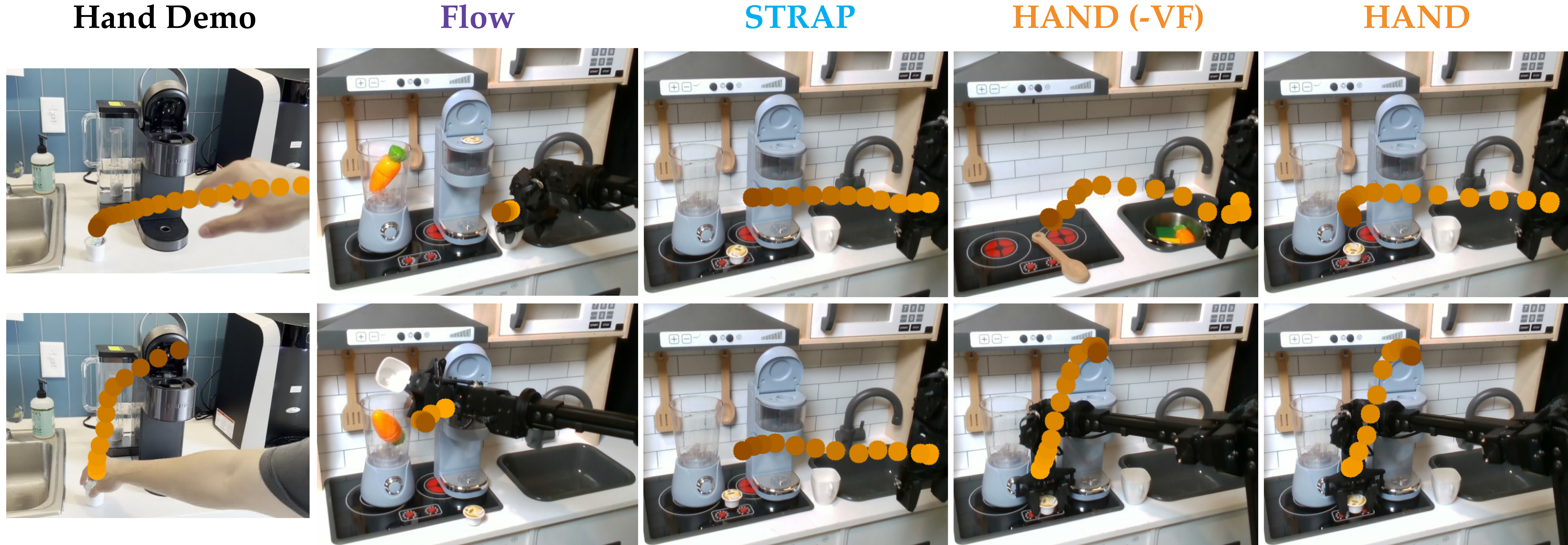}
    \caption{\textbf{Qualitative retrieval results on OOD scene.} 
    We visualize the top sub-trajectory match of \flow , \strap , \method without visual filtering (\handvf), and \hand on two OOD domain demonstrations recorded from an iPhone camera, showing approaching a K-Cup and putting it into the machine. Only \hand's top match is relevant for both hand demonstrations.}
    %We observe that all retrieved trajectories for \method perform the target task while STRAP fails to be robust to the visual gap between the target demo and the play dataset.\todo{fix}}
    \label{fig:qualitative traj analysis}
\end{figure}

\begin{table}[t]
    \centering
    % \vspace{-1em}
    % \addtolength{\tabcolsep}{-1.5pt} % Adjust column separation as needed for this table
    \resizebox{\linewidth}{!}{% <------ Don't forget this %
    \begin{tabular}{lcccccc}
        \toprule
        \textbf{Method} & \textbf{$10^{\circ}$ Horiz.} & \textbf{$20^{\circ}$ Horiz.} & \textbf{$30^{\circ}$ Horiz.} & \textbf{$10^{\circ}$ Vert} & \textbf{$20^{\circ}$ Vert} & \textbf{$30^{\circ}$ Vert} \\
        \midrule
        \flow & 0 / 25 & 2 / 25 & 5 / 25 & 1 / 25 & 0 / 25 & 0 / 25 \\
        \strap & 1 / 25 & 10 / 25 & 13 / 25 & 12 / 25 & 11 / 25 & 11 / 25 \\
        \hand & \textbf{21 / 25} & \textbf{18 / 25} & \textbf{19 / 25} & \textbf{16 / 25} & \textbf{13 / 25} & \textbf{14 / 25} \\
        \bottomrule
    \end{tabular}
    }
    % \addtolength{\tabcolsep}{3.5pt}
    \caption{\textbf{Camera angle robustness results}. \# of relevant retrieved trajectories for the Put Lid on Pot task if we change the camera angle vertically and horizontally by $10^{\circ}$ increments. \hand retrieves $+18\%$ more relevant trajectories compared to \strap even in the extreme case of $30^{\circ}$ shift.}
    \label{tab:camera_angle_results}
\end{table}

\textbf{\ref{q2}: HAND supports hand demonstrations from unseen environments and is robust to camera angle shifts.}
Because \hand retrieves based on \emph{relative hand motions}, it is also effective with hand demonstrations from out-of-distribution (OOD) scenes.
%, provided the camera angle remains similar to relevant trajectories in the play dataset.
To illustrate, we collect hand demonstrations in a new environment using a handheld iPhone camera and a real coffee machine, while retrieving from robot play data recorded in a completely different scene with a toy coffee machine.
In \Cref{fig:qualitative traj analysis}, we show the lowest cost retrieved sub-trajectory of \strap and \flow compared to \hand and a \method ablation without the visual filtering step, \handvf. 
Both of the retrieved trajectories for \strap and \flow, along with the top trajectory for \handvf are irrelevant to the demonstrated task.
For the first task, \strap is able to retrieve the initial reaching motion toward the K-Cup but misses the crucial grasping segment, as it does not leverage motion for retrieval.
Only \hand retrieves relevant robot trajectories for both hand demonstrations because it focuses on the \emph{motion} demonstrated by the human hand after \emph{visual filtering}.
%\hand retrieves more useful trajectories by focusing on the \emph{motion} demonstrated by the hand, after the initial visual filtering step that removes 
%By focusing on the \emph{motion} demonstrated by the human hand after \emph{visual filtering}, \hand retrieves more task-relevant trajectories.

\Cref{tab:camera_angle_results} shows that \textbf{HAND is also robust to shifts in camera angle} for the \task{Put Lid on Pot} task, far more than \flow and \strap.
We measure the number of relevant retrieved trajectories of different methods after vertical and horizontal camera angle shifts of $10^{\circ}$ increments. 
In the most extreme setting of $30^{\circ}$ shift, \hand retrieves $+18\%$ more relevant trajectories compared to \strap.
These camera angle shifts emulate head rotations on humanoid robots or camera movement on mobile manipulators, suggesting that \hand can even work in such settings where camera viewpoint change may occur.% tasks as long as the camera viewpoint is \emph{similar} relative to the arm’s base.

\begin{figure}[t]
    \centering
    \includegraphics[width=\linewidth]{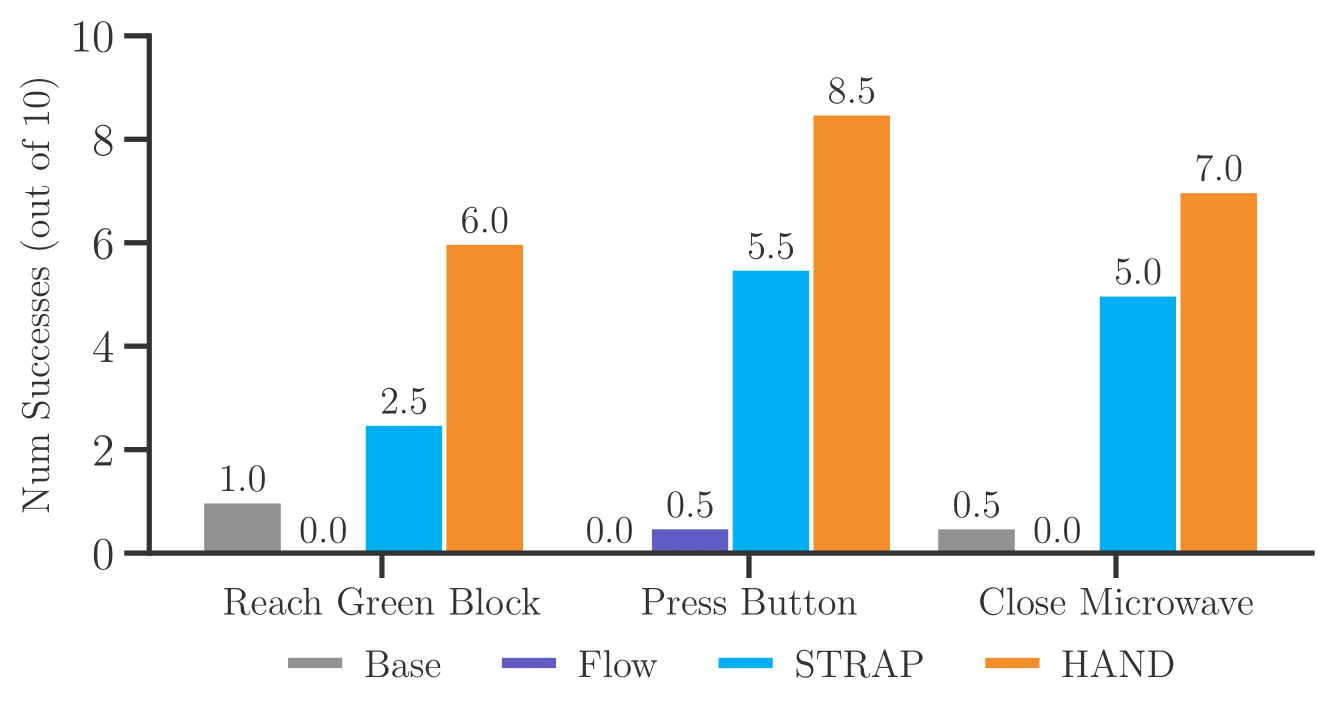}
    \vspace{-20pt}
    \caption{\textbf{Real-Robot Results.} Task completion (including partial completion) out of 10 of $\pi_\text{base}$, \strap, \flow, and \hand.}
    \label{fig:real_robot_hands}
\end{figure}

\begin{table}[t]
    \centering
    % \vspace{-1em}
    \addtolength{\tabcolsep}{-3.5pt}
    \begin{tabular}{lccccc}
        \toprule
        \textbf{Method} & \makecell{\textbf{Slide}\\\textbf{Pot}} & \makecell{\textbf{Put Obj} \\\textbf{in Pot}} & \makecell{\textbf{Put Lid on}\\\textbf{Pot}} & \makecell{\textbf{Turn}\\\textbf{Knob}} & \makecell{\textbf{Long}\\\textbf{Horiz.}} \\
        \midrule
        \langc & 2 & 0 & 0 & 0 & 0 \\
        \clipl & 0 & 4 & 1 & 0 & 0 \\
        \strap & 0 & 0 & 2 & 0 & 0 \\
        \hand w/o Pre-training & 1 & 1 & 1 & 2 & 0 \\
        \handvf & 2 & 0 & 0 & 5 & 0 \\
        \handcw & 2 & 5 & 3 & 5 & 0 \\
        \hand & \textbf{5} & \textbf{6} & \textbf{4} & \textbf{6} & \textbf{3} \\
        \midrule
        \hand + \texttt{Lang Cond} & \textbf{8} & \textbf{7} & \textbf{5} & \textbf{7} & \textbf{5} \\
        \bottomrule
    \end{tabular}
    \caption{\textbf{Long horizon Cook Carrot task results}. We show success rates on each subtask and on the full task execution. Full Task Success is out of 10.}
    \label{tab:long_horizon}
    \addtolength{\tabcolsep}{3.5pt}
\end{table}

\begin{figure*}[t]
    \centering
    \includegraphics[width=\linewidth]{sections/assets/fast_adaptation_study.png}
    \caption{\textbf{Fast Adaptation Study.} We conduct a small-scale user study to demonstrate HAND's ability to learn robot policies in real-time.
    From providing the hand demonstration (Left), to retrieval and fine-tuning a base policy (Middle), to evaluating the policy (Right), we show \hand\ can learn to put a carrot in the blender with 7.5/10 task completion in less than 4 minutes.}
    \label{fig:fast_adaptation}
\end{figure*}

\textbf{\ref{q3}: \method enables efficient policy learning in the real world.} 
% In \Cref{fig:calvin results}, we demonstrate that \hand and ablations outperform baselines in CALVIN, with a \textbf{16\%} average improvement over \flow and \textbf{123\%} over \strap. 
% %highlighting that \hand retrieves trajectories useful for improving downstream performance. 
% In \Cref{fig:calvin results}, we also ablate the use of S-DTW-based loss weighting from \Cref{eq:bc_loss} with \handcw, visual filtering from \Cref{eq:visual_filtering} with \handvf, and ground truth 3D pose information with \handee. 
% \hand outperforms all of these ablations in Move Slider Left. 
% Surprisingly, in this task, \handee\ with priviledged 3D information, even underperforms \handcw. 
% We believe this is because, as \handee\ retrieves trajectories based on an exact match in 3D end-effector pose, the retrieved trajectories have little variability and thus fail to generalize to changes in object placement in the scene.
% In some tasks, we notice that adding visual filtering can negatively impact performance, likely for a similar reason that filtering constrains the diversity of the resulting data subset. 
% However, we demonstrated in the above two paragraphs that visual filtering helps in the real world to retrieve task-relevant trajectories.
%Therefore tuning $M$, the number of visually filtered trajectories, may be important depending on the task/environment.
We evaluate four methods, including ours, across three standard tasks with ten trials each, for a total of \textbf{120 evaluations}. Real-world experiments in \Cref{fig:real_robot_hands} on our three standard tasks
demonstrate that fine-tuning with \hand improves success rates by \textbf{+45\%} over the next best baseline, \strap. In contrast, \flow fails to learn a policy that achieves reasonable success rates in any of the tasks.
% , despite it being the best-performing baseline in CALVIN.
% As we increase $K$, we potentially retrieve more irrelevant trajectories, which can negatively impact the resulting policy.
% Despite visual filtering not always helping in simulation in CALVIN, w
We also report the performance of $\pi_{\text{base}}$, trained on all of $\playdata$ and note that the pre-trained policy struggles to perform the tasks without any task-specific fine-tuning.

We next evaluate eight methods, including several ablations of \hand, across four base tasks and one long-horizon task constructed from these base tasks, with ten trials each for a total of \textbf{400 evaluations}. Results on the more challenging long-horizon tasks (\Cref{tab:long_horizon}) demonstrate that retrieval using \textbf{hand demonstrations outperforms language-based retrieval} (\clipl) by a factor of \textbf{3×} in success rate. Language-based retrieval suffers from the lack of spatial awareness, often retrieving trajectories that are semantically correct but spatially misaligned—similar to \strap—which makes policy fine-tuning more difficult. In contrast, \textbf{directly conditioning on language performs poorly} compared to retrieval (\langc), despite the fact that annotating sub-trajectories with language more than doubles data collection and annotation time.

\textbf{Ablation Study:} We observe that each component of \hand, namely pretraining, visual filtering (VF), and cost weighting (CW), are critical for task performance.
Cost weighting helps bias the resulting policy towards behaviors that are most relevant to the downstream task, and reduces the effect of potentially noisy retrievals that may not directly aid in learning the target task.
Without any of these components, the resulting policy is unable to learn the task.
Only \hand successfully performs the entire long-horizon task reliably. 
We also show that given access to a language-annotated dataset, one could add language-conditioning on top of \hand to further improve the task performance (\hand + \texttt{Lang Cond}).
% We ablate different $K$ values for real robot tasks in \Cref{sec:appendix:additional_real_robot}.
% We hypothesize that retrieving irrelevant trajectories causes the drop in performance for \hand as we continue to increase $K$.

\textbf{\ref{q4}: \method enables real-time, data-efficient policy learning of long-horizon tasks.}
Finally, we performed two small-scale user studies with IRB approval from our institution to demonstrate real-time learning. 
In the first study shown in \Cref{fig:fast_adaptation}, a participant familiar with \method iteratively demonstrated each part of a long-horizon \texttt{Blend Carrot} task and trained a \method policy \textbf{\emph{with over 70\% success rate in under four (4) minutes}} from providing a single hand demonstration to deploying the fine-tuned policy. 
A video of a similar experiment can be found on our website.

% This requires 3 steps: (1) picking up a carrot, (2) putting it in the blender, and (3) pushing the blend button, which were all learned through the retrieved data in such little time.
\begin{table}
    \centering
    \resizebox{\linewidth}{!}{%
        \begin{tabular}{@{}lcc@{}}
        \toprule
        \textbf{Method} & \textbf{User 1 (Minutes)} & \textbf{User 2 (Minutes)} \\ \midrule
        Hand Demonstrations (Min) $\downarrow$  & \textbf{3} & \textbf{2} \\ 
        Robot Demonstrations (Min) $\downarrow$ & 10 & 14 \\ \midrule
        Hand Demonstrations (SR) $\uparrow$ & \textbf{5/10} & \textbf{4/10} \\ 
        Robot Demonstrations (SR) $\uparrow$ & 3/10 & 2.5/10 \\ \bottomrule
        \end{tabular}
    }
    \caption{\textbf{Hand vs Robot Teleoperation}. Time taken and success rates between hand and teleoperated demonstrations.}
    \label{tab:study_two}
\end{table}

\textbf{Hand vs Robot Demonstration Comparison:} In the second study, two users with prior teleoperation experience---but not affiliated with this research---each collected a total of 20 demonstrations: 10 using hand teleoperation and 10 using robot teleoperation, to train the robot for the \texttt{Put K-Cup in Coffee Machine} task.
We employ \hand retrieval for hand-collected demonstrations and \strap retrieval for robot teleoperation demonstrations.
For a direct comparison, we additionally fine-tune \strap with the human-collected teleoperated demonstrations as per \citep{memmel2025strap}.

As reported in \Cref{tab:study_two}, teleoperated demonstrations required over $3 \times$ more time to collect than hand demonstrations.
Remarkably, with just a single hand demonstration per user, we fine-tuned a policy achieving over 40\% task completion compared to \strap which reaches only 25\% using a single \emph{robot teleoperation} demonstration.
Interestingly, we observed that increasing the number of expert demonstrations used for \strap degraded downstream performance likely due to lower quality retrieved trajectories.
These results demonstrate that \hand enables \emph{fast} adaptation to downstream tasks with as few as one (1) easy-to-provide hand demonstration.
%Thus, our results show that hand demos are not only more time-efficient to collect, but also more effective: a single hand demonstration using \hand retrieves more relevant trajectories than \strap using multiple expert demonstrations, indicating superior performance on new downstream tasks.

% \begin{figure}[t]
%     \centering
%     \includegraphics[width=\linewidth]{assets/real_robot_hands}
%     \caption{}
%     \label{fig:real_robot_hands_by_k}
% \end{figure}
%\subfile{sections/06_limitations}
\section{Conclusion and Limitations}
We presented \method, a simple and time-efficient framework for adapting robots to new tasks using easy-to-provide human hand demonstrations.
We demonstrated that \method enables \emph{real-time} task adaptation with a \emph{single} hand demonstration in under four minutes.
% Our results highlight the scalability of \method to train performant real-world, task-specific policies. 
% Finally, we discuss limitations of \method:

\textbf{Extending to 3D paths for retrieval.} While \method\ uses 2D paths for retrieval, one future direction could extend \method\ to estimate the hand trajectory in 3D using foundation depth prediction models.
% Incorporating depth information could provide more fine-grained information about the hand path.
%Furthermore, 2D hand paths do not provide any explicit information about the gripper for retrieval, which could be useful for more dexterous manipulation tasks.
Another direction future work could consider is a mixture of features for improving retrieval for tasks that require more dexterous control, i.e., cloth folding or deformable object manipulation.

% \textbf{Severe Camera Viewpoint Changes.} Although we demonstrated that \method\ is robust for camera viewpoint changes up to $30^\circ$, it would still likely be difficult to retrieve relevant trajectories under severe camera viewpoint between the collected hand demonstration and robot play data.
% Future work could address this issue via the use of 3D information, multiple camera viewpoints, or scene re-rendering with virtual cameras~\citep{goyal2024rvt}.
\textbf{Severe Camera Viewpoint Changes.} Future work could address issues from severe camera viewpoint shifts between the collected hand demonstrations and robot play data via the use of 3D information, multiple camera viewpoints, or scene re-rendering with virtual cameras~\citep{goyal2024rvt}.


%%%%%%%%%%%%%%%%%%%%%%%%%%%%%%%%%%%%%%%%%%%%%%%%%%%%%%%%%%%%%%%%%%%%%%%%%%%%%%%%

\newpage

%%%%%%%%%%%%%%%%%%%%%%%%%%%%%%%%%%%%%%%%%%%%%%%%%%%%%%%%%%%%%%%%%%%%%%%%%%%%%%%%

%%%%%%%%%%%%%%%%%%%%%%%%%%%%%%%%%%%%%%%%%%%%%%%%%%%%%%%%%%%%%%%%%%%%%%%%%%%%%%%%
%\section*{APPENDIX}
%
%
%\section*{ACKNOWLEDGMENT}

%%%%%%%%%%%%%%%%%%%%%%%%%%%%%%%%%%%%%%%%%%%%%%%%%%%%%%%%%%%%%%%%%%%%%%%%%%%%%%%%
\balance
\printbibliography

@inproceedings{zhao23aloha,
  author       = {Tony Z. Zhao and
                  Vikash Kumar and
                  Sergey Levine and
                  Chelsea Finn},
  editor       = {Kostas E. Bekris and
                  Kris Hauser and
                  Sylvia L. Herbert and
                  Jingjin Yu},
  title        = {Learning Fine-Grained Bimanual Manipulation with Low-Cost Hardware},
  booktitle    = {Robotics: Science and Systems},
  year         = {2023}
}

@inproceedings{shridhar2023perceiver,
  title={Perceiver-actor: A multi-task transformer for robotic manipulation},
  author={Shridhar, Mohit and Manuelli, Lucas and Fox, Dieter},
  booktitle={Conference on Robot Learning},
  year={2023},
}

@article{goyal2024rvt,
  author    = {Goyal, Ankit and Blukis, Valts and Xu, Jie and Guo, Yijie and Chao, Yu-Wei and Fox, Dieter},
  title     = {RVT2: Learning Precise Manipulation from Few Demonstrations},
  journal   = {Robotics: Science and Systems},
  year      = {2024},
}

@article{kim2024openvla,
  title={OpenVLA: An Open-Source Vision-Language-Action Model},
  author={Kim, Moo Jin and Pertsch, Karl and Karamcheti, Siddharth and Xiao, Ted and Balakrishna, Ashwin and Nair, Suraj and Rafailov, Rafael and Foster, Ethan and Lam, Grace and Sanketi, Pannag and others},
  journal={arXiv preprint arXiv:2406.09246},
  year={2024}
}

@inproceedings{walke2023bridgedata,
    title={BridgeData V2: A Dataset for Robot Learning at Scale},
    author={Walke, Homer and Black, Kevin and Lee, Abraham and Kim, Moo Jin and Du, Max and Zheng, Chongyi and Zhao, Tony and Hansen-Estruch, Philippe and Vuong, Quan and He, Andre and Myers, Vivek and Fang, Kuan and Finn, Chelsea and Levine, Sergey},
    booktitle={Conference on Robot Learning},
    year={2023}
}

@article{team2024octo,
  title={Octo: An open-source generalist robot policy},
  author={Team, Octo Model and Ghosh, Dibya and Walke, Homer and Pertsch, Karl and Black, Kevin and Mees, Oier and Dasari, Sudeep and Hejna, Joey and Kreiman, Tobias and Xu, Charles and others},
  journal={arXiv preprint arXiv:2405.12213},
  year={2024}
}

@inproceedings{xu2024flow,
  title={Flow as the Cross-domain Manipulation Interface},
  author={Xu, Mengda and Xu, Zhenjia and Xu, Yinghao and Chi, Cheng and Wetzstein, Gordon and Veloso, Manuela and Song, Shuran},
  booktitle={Conference on Robot Learning},
  year={2024}
}

@article{black2024pi0,
  title={$pi\_0$: A Vision-Language-Action Flow Model for General Robot Control},
  author={Black, Kevin and Brown, Noah and Driess, Danny and Esmail, Adnan and Equi, Michael and Finn, Chelsea and Fusai, Niccolo and Groom, Lachy and Hausman, Karol and Ichter, Brian and others},
  journal={arXiv preprint arXiv:2410.24164},
  year={2024}
}

@article{kuang2024ram,
  title={Ram: Retrieval-based affordance transfer for generalizable zero-shot robotic manipulation},
  author={Kuang, Yuxuan and Ye, Junjie and Geng, Haoran and Mao, Jiageng and Deng, Congyue and Guibas, Leonidas and Wang, He and Wang, Yue},
  journal={Conference on Robot Learning},
  year={2024}
}

@inproceedings{karaev2025cotracker,
  title={Cotracker: It is better to track together},
  author={Karaev, Nikita and Rocco, Ignacio and Graham, Benjamin and Neverova, Natalia and Vedaldi, Andrea and Rupprecht, Christian},
  booktitle={European Conference on Computer Vision},
  year={2025},
}

@inproceedings{
li2025hamster,
title={{HAMSTER}: Hierarchical Action Models for Open-World Robot Manipulation},
author={Yi Li and Yuquan Deng and Jesse Zhang and Joel Jang and Marius Memmel and Caelan Reed Garrett and Fabio Ramos and Dieter Fox and Anqi Li and Abhishek Gupta and Ankit Goyal},
booktitle={International Conference on Learning Representations},
year={2025},
}

@misc{lepert2025phantomtrainingrobotsrobots,
        title={Phantom: Training Robots Without Robots Using Only Human Videos}, 
        author={Marion Lepert and Jiaying Fang and Jeannette Bohg},
        year={2025},
        eprint={2503.00779},
        archivePrefix={arXiv},
        primaryClass={cs.RO}
  }

@inproceedings{lin2024flowretrieval,
  title={FlowRetrieval: Flow-Guided Data Retrieval for Few-Shot Imitation Learning},
  author={Lin, Li-Heng and Cui, Yuchen and Xie, Amber and Hua, Tianyu and Sadigh, Dorsa},
  booktitle={Conference on Robot Learning},
  year={2024}
}

@inproceedings{
memmel2025strap,
title={{STRAP}: Robot Sub-Trajectory Retrieval for Augmented Policy Learning},
author={Marius Memmel and Jacob Berg and Bingqing Chen and Abhishek Gupta and Jonathan Francis},
booktitle={International Conference on Learning Representations},
year={2025}
}

@inproceedings{du2023behaviorretrievalfewshotimitation,
      title={Behavior Retrieval: Few-Shot Imitation Learning by Querying Unlabeled Datasets}, 
      author={Maximilian Du and Suraj Nair and Dorsa Sadigh and Chelsea Finn},
      year={2023},  
      booktitle={Robotics: Science and Systems}
}

@inproceedings{nasiriany2022sailo,
  title={Learning and Retrieval from Prior Data for Skill-based Imitation Learning},
  author={Soroush Nasiriany and Tian Gao and Ajay Mandlekar and Yuke Zhu},
  booktitle={Conference on Robot Learning},
  year={2022}
}

@article{haldar2025point,
    title={Point Policy: Unifying Observations and Actions with Key Points for Robot Manipulation},
    author={Haldar, Siddhant and Pinto, Lerrel},
    journal={arXiv preprint arXiv:2502.20391},
    year={2025}
  }

@article{kim2023giving,
  title={Giving Robots a Hand: Learning Generalizable Manipulation with Eye-in-Hand Human Video Demonstrations},
  author={Kim, Moo Jin and Wu, Jiajun and Finn, Chelsea},
  journal={CoRR},
  year={2023}
}

@misc{papagiannis2024rxretrievalexecutioneveryday,
      title={R+X: Retrieval and Execution from Everyday Human Videos}, 
      author={Georgios Papagiannis and Norman Di Palo and Pietro Vitiello and Edward Johns},
      year={2024},
      eprint={2407.12957},
      archivePrefix={arXiv},
      primaryClass={cs.RO}
}

@misc{geminiroboticsteam2025geminiroboticsbringingai,
      title={Gemini Robotics: Bringing AI into the Physical World}, 
      author={Gemini Robotics Team and others},
      year={2025},
      eprint={2503.20020},
      archivePrefix={arXiv},
      primaryClass={cs.RO},
      url={https://arxiv.org/abs/2503.20020}, 
}

@article{mees2022calvin,
author = {Oier Mees and Lukas Hermann and Erick Rosete-Beas and Wolfram Burgard},
title = {CALVIN: A Benchmark for Language-Conditioned Policy Learning for Long-Horizon Robot Manipulation Tasks},
journal={Robotics and Automation Letters (RA-L)},
year={2022}
}

@InProceedings{lynch2020playdata,
  title = 	 {Learning Latent Plans from Play},
  author =       {Lynch, Corey and Khansari, Mohi and Xiao, Ted and Kumar, Vikash and Tompson, Jonathan and Levine, Sergey and Sermanet, Pierre},
  booktitle = 	 {Conference on Robot Learning},
  year = 	 {2020},
}

@inproceedings{young2022playful,
  title={Playful interactions for representation learning},
  author={Young, Sarah and Pari, Jyothish and Abbeel, Pieter and Pinto, Lerrel},
  booktitle={International Conference on Intelligent Robots and Systems},
  year={2022},
  organization={IEEE}
}

@article{molmo2024,
  title={Molmo and PixMo: Open Weights and Open Data for State-of-the-Art Multimodal Models},
  author={Matt Deitke and others},
  journal={arXiv preprint arXiv:2409.17146},
  year={2024}
}

@inproceedings{karaev23cotracker,
  title     = {CoTracker: It is Better to Track Together},
  author    = {Nikita Karaev and Ignacio Rocco and Benjamin Graham and Natalia Neverova and Andrea Vedaldi and Christian Rupprecht},
  booktitle = {Proc. {ECCV}},
  year      = {2024}
}

@inproceedings{
  zhang2024extract,
  title={{EXTRACT}: Efficient Policy Learning by Extracting Transferrable Robot Skills from Offline Data},
  author={Jesse Zhang and Minho Heo and Zuxin Liu and Erdem Biyik and Joseph J Lim and Yao Liu and Rasool Fakoor},
  booktitle={Conference on Robot Learning},
  year={2024}
}

@misc{zhao2024alohaunleashedsimplerecipe,
      title={ALOHA Unleashed: A Simple Recipe for Robot Dexterity}, 
      author={Tony Z. Zhao and Jonathan Tompson and Danny Driess and Pete Florence and Kamyar Ghasemipour and Chelsea Finn and Ayzaan Wahid},
      year={2024},
      eprint={2410.13126},
      archivePrefix={arXiv},
      primaryClass={cs.RO}
}

@article{yuan2024general,
        title={General Flow as Foundation Affordance for Scalable Robot Learning},
        author={Yuan, Chengbo and Wen, Chuan and Zhang, Tong and Gao, Yang},
        journal={Conference on Robot Learning},
        year={2024}
      }

@misc{oquab2024dinov2learningrobustvisual,
      title={DINOv2: Learning Robust Visual Features without Supervision}, 
      author={Maxime Oquab and Timothée Darcet and Théo Moutakanni and Huy Vo and Marc Szafraniec and Vasil Khalidov and Pierre Fernandez and Daniel Haziza and Francisco Massa and Alaaeldin El-Nouby and Mahmoud Assran and Nicolas Ballas and Wojciech Galuba and Russell Howes and Po-Yao Huang and Shang-Wen Li and Ishan Misra and Michael Rabbat and Vasu Sharma and Gabriel Synnaeve and Hu Xu and Hervé Jegou and Julien Mairal and Patrick Labatut and Armand Joulin and Piotr Bojanowski},
      year={2024},
      eprint={2304.07193},
      archivePrefix={arXiv},
      primaryClass={cs.CV},
}

@inproceedings{bahl2023affordances,
              title={Affordances from Human Videos as a Versatile Representation for Robotics},
              author={Bahl, Shikhar and Mendonca, Russell and Chen, Lili and Jain, Unnat and Pathak, Deepak},
              journal={Conference on Computer Vision and Pattern Recognition},
              year={2023}
            }

@inproceedings{
    liu2024tail,
    title={{TAIL}: Task-specific Adapters for Imitation Learning with Large Pretrained Models},
    author={Zuxin Liu and Jesse Zhang and Kavosh Asadi and Yao Liu and Ding Zhao and Shoham Sabach and Rasool Fakoor},
    booktitle={International Conference on Learning Representations},
    year={2024},
}

@article{hu2022lora,
  title={Lora: Low-rank adaptation of large language models.},
  author={Hu, Edward J and Shen, Yelong and Wallis, Phillip and Allen-Zhu, Zeyuan and Li, Yuanzhi and Wang, Shean and Wang, Lu and Chen, Weizhu and others},
  journal={International Conference on Learning Representations},
  year={2022}
}

@inproceedings{liang2022transformer,
  title={Transformer adapters for robot learning},
  author={Liang, Anthony and Singh, Ishika and Pertsch, Karl and Thomason, Jesse},
  booktitle={CoRL 2022 Workshop on Pre-training Robot Learning},
  year={2022}
}

@article{zhao2023learning,
  title={Learning fine-grained bimanual manipulation with low-cost hardware},
  author={Zhao, Tony Z and Kumar, Vikash and Levine, Sergey and Finn, Chelsea},
  journal={arXiv preprint arXiv:2304.13705},
  year={2023}
}

@inproceedings{
sridhar2025regent,
title={{REGENT}: A Retrieval-Augmented Generalist Agent That Can Act In-Context in New Environments},
author={Kaustubh Sridhar and Souradeep Dutta and Dinesh Jayaraman and Insup Lee},
booktitle={International Conference on Learning Representations},
year={2025}
}

@article{peng2019advantage,
  title={Advantage-weighted regression: Simple and scalable off-policy reinforcement learning},
  author={Peng, Xue Bin and Kumar, Aviral and Zhang, Grace and Levine, Sergey},
  journal={arXiv preprint arXiv:1910.00177},
  year={2019}
}

@article{haldar2024baku,
  title={Baku: An efficient transformer for multi-task policy learning},
  author={Haldar, Siddhant and Peng, Zhuoran and Pinto, Lerrel},
  journal={Neural Information Processing Systems},
  year={2024}
}

@book{muller2021fundamentals,
  title={Fundamentals of music processing: Using Python and Jupyter notebooks},
  author={M{\"u}ller, Meinard},
  volume={2},
  year={2021},
  publisher={Springer}
}

@inproceedings{xu2022gmflow,
  title={Gmflow: Learning optical flow via global matching},
  author={Xu, Haofei and Zhang, Jing and Cai, Jianfei and Rezatofighi, Hamid and Tao, Dacheng},
  booktitle={Conference on Computer Vision and Pattern Recognition},
  year={2022}
}

@misc{kareer2024egomimicscalingimitationlearning,
      title={EgoMimic: Scaling Imitation Learning via Egocentric Video}, 
      author={Simar Kareer and Dhruv Patel and Ryan Punamiya and Pranay Mathur and Shuo Cheng and Chen Wang and Judy Hoffman and Danfei Xu},
      year={2024},
      eprint={2410.24221},
      archivePrefix={arXiv},
      primaryClass={cs.RO}
    }

@misc{li2025trainrobotsimpactdemonstration,
      title={How to Train Your Robots? The Impact of Demonstration Modality on Imitation Learning}, 
      author={Haozhuo Li and Yuchen Cui and Dorsa Sadigh},
      year={2025},
}

@Article{hand_vs_robto_mdpi_2025,
AUTHOR = {Xie, Jianan and Xu, Zhen and Zeng, Jiayu and Gao, Yuyang and Hashimoto, Kenji},
TITLE = {Human–Robot Interaction Using Dynamic Hand Gesture for Teleoperation of Quadruped Robots with a Robotic Arm},
JOURNAL = {Electronics},
YEAR = {2025}
}

@inproceedings{kedia2025oneshotimitationmismatchedexecution,
  title={One-Shot Imitation under Mismatched Execution}, 
  author={Kushal Kedia and Prithwish Dan and Angela Chao and Maximus Adrian Pace and Sanjiban Choudhury},
  year={2025},
  booktitle={International Conference on Robotics and Automation (ICRA)}
}

@inproceedings{
lin2025data,
title={Data Scaling Laws in Imitation Learning for Robotic Manipulation},
author={Fanqi Lin and Yingdong Hu and Pingyue Sheng and Chuan Wen and Jiacheng You and Yang Gao},
booktitle={The Thirteenth International Conference on Learning Representations},
year={2025},
}

@inproceedings{xie2025data,
  title={Data Retrieval with Importance Weights for Few-Shot Imitation Learning},
  author={Xie, Amber and Chand, Rahul and Sadigh, Dorsa and Hejna, Joey},
  booktitle={9th Annual Conference on Robot Learning},
  year={2025},
  url={https://arxiv.org/abs/2509.01657}
}

@inproceedings{sridhar2025ricladdingincontextadaptability,
      title={RICL: Adding In-Context Adaptability to Pre-Trained Vision-Language-Action Models}, 
      author={Kaustubh Sridhar and Souradeep Dutta and Dinesh Jayaraman and Insup Lee},
      booktitle={Conference on Robot Learning (CoRL)},
      year={2025},
      organization={PMLR}
}
%\bibliographystyle{IEEEtran}
%\bibliography{references.bib}

\end{document}